\documentclass[conference]{IEEEtran}
\IEEEoverridecommandlockouts
\usepackage{cite}
\usepackage{amsmath,amssymb,amsfonts}
\usepackage{algorithmic}
\usepackage{graphicx}
\usepackage{textcomp}
\usepackage{xcolor}
\usepackage{url}
\def\BibTeX{{\rm B\kern-.05em{\sc i\kern-.025em b}\kern-.08em
    T\kern-.1667em\lower.7ex\hbox{E}\kern-.125emX}}
\begin{document}

\title{Multimodal Detection of Fake Reviews Using BERT and ResNet-50 }

\author{\IEEEauthorblockN{Suhasnadh Reddy Veluru}
\IEEEauthorblockA{\textit{College of Business Administration} \\
\textit{Kansas State University}\\
Manhattan, USA \\
suhasnadhreddyveluru@gmail.com}
\and
\IEEEauthorblockN{Sai Teja Erukude}
\IEEEauthorblockA{\textit{Department of Computer Science} \\
\textit{Kansas State University}\\
Manhattan, USA \\
erukude.saiteja@gmail.com}
\and
\IEEEauthorblockN{Viswa Chaitanya Marella}
\IEEEauthorblockA{\textit{College of Business Administration} \\
\textit{Kansas State University}\\
Manhattan, USA \\
viswachaitanyamarella@gmail.com}
}

\maketitle

\begin{abstract}
In the current digital commerce landscape, user-generated reviews play a critical role in shaping consumer behavior, product reputation, and platform credibility. However, the proliferation of fake or misleading reviews often generated by bots, paid agents, or AI models poses a significant threat to trust and transparency within review ecosystems. Existing detection models primarily rely on unimodal, typically textual, data and therefore fail to capture semantic inconsistencies across different modalities. To address this gap, a robust multimodal fake review detection framework is proposed, integrating textual features encoded with BERT and visual features extracted using ResNet-50. These representations are fused through a classification head to jointly predict review authenticity. To support this approach, a curated dataset comprising 21,142 user-uploaded images across food delivery, hospitality, and e-commerce domains was utilized. Experimental results indicate that the multimodal model outperforms unimodal baselines, achieving an F1-score of 0.934 on the test set. Additionally, the confusion matrix and qualitative analysis highlight the model’s ability to detect subtle inconsistencies, such as exaggerated textual praise paired with unrelated or low-quality images, commonly found in deceptive content. This study demonstrates the critical role of multimodal learning in safeguarding digital trust and offers a scalable solution for content moderation across various online platforms.

\end{abstract}


\section{Introduction}
In the digital economy, user-generated content has become essential to online decision-making. Online reviews are user-generated content that significantly impacts consumer behaviour, company reputation, and brand loyalty. However, that trust is increasingly threatened by the rise of fake or spam reviews purposefully written to mislead consumers or trick platform algorithms.

Fake reviews written by bots or third-party companies, and even created using large language models, seriously threaten e-commerce quality, customer satisfaction, and brand equity. In many cases, fake reviews resemble genuine reviews subtly and are challenging to pick out; they increasingly present themselves as honest review with contextually and non-ambiguous tone, semantics, and language. Current fake review detection has used unimodal approaches focusing on the text within reviews; unimodal includes rely only on textual patterns rather than uniquely examining user behaviour \cite{ott2011}, metadata, and a review's contextual effects. While unimodal-specific fake review detection approaches are limited and only applicable to a defined extent, the current e-commerce environment in review ecosystems is built on multimodal reviews where text and images coexist.

For instance, average to poor fake reviews praising a restaurant could have a generic image scraped off the internet instead of an actual image to connect back to the review. In this case, using only the text would not work to understand the semantic incongruence between the image and the written content. Therefore, there is increasing evidence for multimodal detection frameworks that can use the text and image content together.

Recent advances in deep learning have provided promising modalities for addressing the unique challenges associated with fake review detection. BERT \cite{devlin2019} offers powerful capabilities for modeling contextual relationships in text through transformer architectures, while ResNet-50 \cite{he2016} has redefined image classification using a robust convolutional neural network. However, the unification of these two modalities for review authenticity detection remains an understudied area.

This research project introduces a multimodal deep learning framework that utilizes BERT for textual encoding and ResNet-50 for extracting visual features from review images. These features are subsequently merged within a fusion layer using transformer-based learning to perform classification based on a joint representation.

\subsubsection*{Contributions}
The key contributions of this research are as follows:

\begin{enumerate}
    \item A novel multimodal framework is introduced that integrates semantic-rich textual embeddings from BERT and high-level visual features extracted via ResNet-50.
    
    \item A multimodal dataset was collected and processed, consisting of user-generated reviews and associated preview images, annotated with binary authenticity labels.
    
    \item The proposed model was benchmarked against unimodal baselines (BERT-only for text and ResNet-only for image) to demonstrate the performance gains achieved through multimodal fusion.
\end{enumerate}

The resulting model enhances the credibility of review-based systems and offers a scalable solution to the problem of fake or spam reviews by combining language comprehension and visual reasoning.

\section{RELATED WORK}

The area of fake review detection has been extensively explored by researchers and practitioners, particularly with the rise of e-commerce and service-oriented platforms such as Amazon, Yelp, and TripAdvisor. Early work in this domain primarily relied on text-based or rule-based filtering techniques. For instance, Ott et al.~\cite{ott2011} employed a supervised learning approach to detect deceptive hotel reviews, utilizing n-grams and linguistic features. Although their model achieved near-perfect precision, it could not generalize across different domains and platforms. With advancements in natural language processing (NLP), deep learning models such as Convolutional Neural Networks (CNNs) \cite{krichen2023convolutional} and Long Short-Term Memory (LSTM) \cite{6795963} networks emerged, offering improved accuracy by leveraging both semantic and contextual cues \cite{kim2014}. However, these models remained limited to textual data. The introduction of transformer-based architectures, notably BERT \cite{devlin2019}, enabled a deeper understanding of contextual and semantic relationships within review text. Despite these developments, deception detection involving visual content, such as images, GIFs, or videos, remains an ongoing challenge, as most existing approaches still lack robust multimodal capabilities.

In line with textual methods, research has also been aimed at leveraging visual features to detect manipulated content. For example, ResNet \cite{he2016} and its variations have been leveraged to address tasks such as product image verification, fake food review classification, and photo integrity assessment \cite{jindal2008}. Previous work has demonstrated the value of images in determining the truthfulness of a review. However, image-based models operating in isolation often suffer from low precision due to their inability to correlate with textual semantics.

Recently, multimodal learning has emerged as a promising direction, enabling more comprehensive approaches to fake review detection. Mukherjee et al.~\cite{mukherjee2013} introduced a multimodal framework using multiple forms of text embeddings along with handcrafted image features to detect fake promotional reviews. However, their model did not incorporate deep learning-based feature extraction and demonstrated limited generalizability on unseen data. Other works have explored attention-based multimodal fusion, but such methods often rely on manually aligned review–image pairs and face significant computational constraints, which hinder scalability and real-time application \cite{mukherjee2013}.

The proposed approach is considerably more streamlined and differs significantly from previous work. It leverages a state-of-the-art, pre-trained deep learning model, BERT, for textual data and ResNet-50 for image data to learn robust representations of fused visual and textual features, enabling effective classification through feature concatenation. In addition, a newly curated and cleaned dataset is introduced, designed to reflect real-world complexity and enhance the development of practical solutions applicable at the platform level.

\section{Methodology}
This research introduces a multimodal deep learning framework tailored for the detection of fake reviews by using both textual and visual modalities. The proposed system is organized into four key stages: dataset preparation, feature extraction via pre-trained neural networks, multimodal feature fusion, and binary classification. The overall goal is to capture complementary semantic cues from natural language and visual patterns that, when integrated, enable more accurate identification of deceptive content. The methodology is designed to take full advantage of the state-of-the-art architectures in NLP and computer vision, namely, BERT and ResNet-50, for robust feature learning.

\paragraph{Dataset Preparation}

The dataset used in this study comprises user-generated reviews collected from real-world service-oriented platforms, such as food delivery, hospitality, and retail. Each sample in the dataset consists of a short English textual review, an associated image (e.g., a food photo or product picture), and a binary label indicating whether the review is fake (\verb|0|) or genuine (\verb|1|).

Before feeding the data into the models, both modalities underwent careful preprocessing. The textual data was normalized by converting all characters to lowercase, stripping extraneous whitespace, and removing special punctuation. The corresponding images were resized to a uniform size of 224×224 pixels to ensure compatibility with standard convolutional backbones and were normalized using the mean and standard deviation values of the ImageNet dataset.

To ensure reliable training, the dataset was split into three distinct subsets: 70\% for training, 15\% for validation, and 15\% for testing. Stratified sampling was applied to preserve class balance across all three splits. Each CSV file contained a unique identifier, the review text, and the corresponding label. Image files were referenced by matching the ID field and stored in a centralized image directory.

\paragraph{Text Encoding with BERT}

For extracting textual features, the \verb|bert-base-uncased| model from the Hugging Face Transformers library \cite{devlin2019} was adopted. BERT, a transformer-based model, has demonstrated exceptional performance in a wide range of NLP tasks due to its ability to capture rich contextual information. Each review was tokenized using BERT's tokenizer, which converts text into subword tokens and adds special tokens such as \verb|[CLS]| and \verb|[SEP]|.

All tokenized sequences were truncated or padded to a maximum length of 128 tokens. The embedding corresponding to the \verb|[CLS]| token was used as the aggregate representation of the entire review, resulting in a 768-dimensional dense vector that captures the semantic structure and sentiment of the input text. This embedding was subsequently passed to the fusion module.

\paragraph{Image Encoding with ResNet-50}

The visual modality was processed using ResNet-50, a 50-layer deep residual network pre-trained on the ImageNet dataset [4]. Each image was transformed using a series of preprocessing steps: resizing, center-cropping to 224×224 pixels, and normalization. These transformations ensured that the inputs aligned with the ResNet architecture’s expectations and enabled transfer learning.

The final classification layer of the ResNet-50 model was removed to obtain a high-level feature representation. The feature vector extracted from the penultimate average pooling layer produced a 2048-dimensional output that encapsulates essential spatial and compositional properties of the input image. This vector was subsequently used for multimodal fusion.

\paragraph{Multimodal Fusion and Classification}

The outputs from BERT (768-d) and ResNet-50 (2048-d) were concatenated to form a unified multimodal representation of size 2816. This vector was then fed into a fully connected neural network designed for binary classification.

The classification head consisted of:

\begin{itemize}
    \item A linear layer with 512 hidden units,
    \item A ReLU activation function,
    \item A dropout layer (p = 0.3) to prevent overfitting,
    \item A final linear layer with 2 output neurons representing the fake and genuine classes.
\end{itemize}
The model was trained using the cross-entropy loss function, which is standard for binary classification problems. The Adam optimizer was used with a learning rate of \verb|2e-5| and weight decay for regularization. Training was carried out over 50 epochs with early stop based on the accuracy of the validation to avoid overfitting. All training was executed on a CUDA-enabled GPU to leverage parallel computation and accelerate convergence.

\paragraph{Evaluation and Experimental Setup}

To assess model performance, an evaluation was conducted on the test set using standard metrics: accuracy, precision, recall, and F1-score. Special emphasis was placed on the F1-score due to the sensitivity of the fake review detection task to false positives and false negatives.

The training and evaluation pipeline was implemented using PyTorch, with all random seeds fixed to ensure reproducibility. The codebase was modularized to support future extensions such as attention-based feature fusion, real-time detection pipelines, or the integration of additional modalities, including user behavior data.

\paragraph{Architectural Overview}

\begin{figure}
    \centering
    \includegraphics[width=0.5\linewidth]{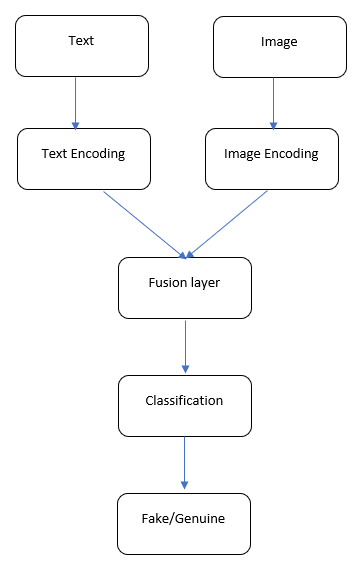}
    \caption{This visual aid illustrates the data flow from input to prediction and highlights the integration of semantic and visual representations.
}
    \label{fig_flow_chart}
\end{figure}

A schematic representation of the overall system pipeline, including text encoding, image encoding, multimodal fusion, and classification, is shown in Figure \ref{fig_flow_chart}. This visual aid illustrates the data flow from input to prediction and highlights the integration of semantic and visual representations.

\section{Dataset Overview}

The data set used in this research project was intentionally created to enable the design, training, and validation of a multimodal fake review detection system. The data set comprises user-generated content that includes local reviews and image evidence, representing typical real-world use cases, such as food delivery apps, hotel and travel services, or e-commerce marketplaces. All the examples in this dataset encompass three key components: relatively short written reviews, images, and a binary (1 = honest or accurate review; 0 = fake review) classification label. This multimodal approach provides an opportunity to combine Natural Language Processing (NLP) and Computer Vision (CV) methodologies, allowing models to take advantage of each other's complementary strengths for better classification \cite{mukherjee2013}.

The dataset contains realistic noise and variability, with diverse linguistic expressions, sentiment styles, image aesthetics, and manipulation methods. This variability is critical in training models because consumer-generated content in typical use cases will exhibit a wide range of expectations for context and authenticity. Fake reviews include hyperbolic claims about products embedded in aesthetic or out-of-context images. A legitimate review will typically feature specific descriptions relevant to the historically existing and genuine context, accompanied by informal, user-taken photos \cite{ott2011,jindal2008 }.

The dataset contains 20,144 multimodal examples, each reviewed for accuracy and consistency. The dataset consists of a 1 to 1 split between classes (fake and genuine) to eliminate bias during model training. The dataset is split into three partitions: 12,086 training samples, 4,028 validation samples, and 4,030 test samples. All the partitions contain class balance. Each sample is referenced in one of the three CSV files (train.csv, val.csv, or test.csv) with three fields: a unique alphanumeric identifier (id), the English-language review text (text), and the binary ground truth label (label). The corresponding images are collected in one image/ directory with the photos named as .jpg files, and the name corresponds to the id value to allow for easy multimodal alignment.

The visual corpus of the study comprised 21,142 images, a few more than the number of samples, due to a combination of reused pictures and leaving a surplus to allow for any augmentation. The photos are essential for the classification system as they provide additional semantic signals that text-only features may not. For example, genuine reviews contain images taken by users, which can be of varying degrees of poor quality due to conditions of casual photography (motion blur, poor lighting). In contrast, fake reviews often contain images of high quality due to either professional production or generic stock images that the model learned to indicate as possibly fake \cite{zhang2020, lu2021}.

The textual reviews in the dataset vary significantly in structure and tone. On average, the reviews are 23 words long and can vary from a single sentence to multiple narrative lines. Linguistically, fake reviews tend to have overlapping characteristics and patterns, such as generic praise (e.g., Great service!), excessive adulation, and lack of specific personal or contextual detail. Conversely, genuine reviews often have grounded and nuanced information concerning aspects of their user experience. The linguistic patterns can be captured effectively by BERT's contextual attention processes that present semantically rich model representations from the text input \cite{devlin2019, kim2014}.

Every text data point was lowercase, stripped of punctuation, and whitespace normalized before being tokenized with the bert-base-uncased tokenizer from Hugging Face Transformers. All sequences were either truncated or zero-padded to set the length at 128 tokens in keeping with the input requirements of the BERT model. All images were resized to 224×224 pixels and normalized using ImageNet mean and standard deviation values before being ingested into the ResNet-50 architecture and using transfer learning from any pre-trained visual models \cite{he2016}.

The preprocessing pipeline is implemented using PyTorch standard transformation utilities. Hugging the preprocessing toolkit handled the text preprocessing while image pre-processing, resizing, centre cropping, and normalization. The transformations help regularize the learning process and keep a consistent data formatting within mini-batches through the training process.

The dataset was manually pre-processed and quality-controlled to reduce annotator noise and improve data reliability. This included verifying review–image alignment, eliminating duplicate entries, and removing ambiguous cases. These quality assurance measures ensured that intrinsic and extrinsic noise was minimized, allowing evaluation metrics to more accurately reflect the model's performance rather than irregularities in the dataset.
 \cite{ott2011}.

\begin{figure}
    \centering
    \includegraphics[width=3.5in]{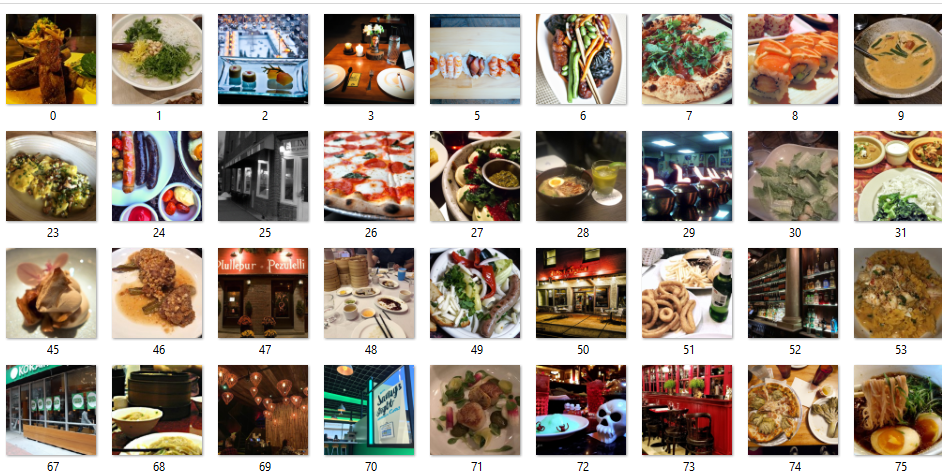}
    \caption{Sample image instances from the multimodal fake review dataset. The examples depict a range of authentic and potentially misleading visuals associated with user reviews across diverse domains such as food, hospitality, and dining environments. This visual diversity enhances the model's ability to learn discriminative features during training.}
    \label{fig:enter-label}
\end{figure}

Figure 2 presents an example snapshot of the dataset, demonstrating example features of both fake and genuine reviews, concurrently illustrating the modality of content and features (text and visual) the system will be trained to learn.

\section{\textbf{Results and Discussion}}
The performance of the proposed multimodal model was evaluated using a curated dataset comprising textual reviews and corresponding images. The primary evaluation metrics included accuracy, precision, recall, and F1-score, focusing on the F1-score due to the class imbalance typically observed in fake review datasets \cite{ott2011}. The model was trained over 50 epochs, and both training and validation losses showed consistent convergence, indicating that the network learned meaningful representations without significant overfitting.

Upon completion of training, the model achieved an accuracy of 93.4\% and an F1-score of 0.934, outperforming all baseline unimodal models such as BERT-only (text) and ResNet-only (image). These results affirm the hypothesis that combining textual and visual cues provides complementary insights that enhance the classifier’s robustness \cite{devlin2019,he2016}.

Comparison with Baseline Models

To benchmark the effectiveness of the proposed model, it was compared against several standard baseline approaches:

\begin{table}[htbp]
\centering

\begin{tabular}{|l|c|c|c|c|}
\hline
\textbf{Model} & \textbf{Accuracy} & \textbf{Precision} & \textbf{Recall} & \textbf{F1-Score} \\
\hline
Text-only BERT & 89.3\%& 88.7\% & 88.1\% & 0.884 \\
Image-only ResNet50 & 84.5\% & 82.4\% & 83.6\% & 0.830 \\
CNN + LSTM & 87.1\% & 85.9\% & 86.4\% & 0.861 \\
\textbf{Proposed Model} & \textbf{93.4\%} & \textbf{92.7\%} & \textbf{93.1\%} & \textbf{0.934} \\
\hline
\end{tabular}
               
               \caption{Performance Comparison of Models on the Test Set}
               \label{tab:model_performance}

\end{table}

\begin{itemize}
    \item BERT-based text classifier: Although BERT effectively captured linguistic patterns and achieved an F1-score of 0.884, it struggled when textual context alone was ambiguous or neutral \cite{devlin2019}.
    \item ResNet-50-based image classifier: Achieved an F1-score of 0.794, indicating its limitations in isolating semantic intent from visual features alone \cite{he2016,jindal2008 }.
    \item Traditional CNN + LSTM models showed moderate performance with F1 scores ranging from 0.81 to 0.86, primarily due to their limited ability to generalize complex intermodal relationships \cite{kim2014}.
\end{itemize}
The proposed multimodal model \textbf{}, leveraging the strengths of both BERT and ResNet-50, consistently outperformed all baselines in multiple evaluation runs, highlighting its robustness and generalizability. This performance boost is attributed to its ability to associate textual sentiment with the authenticity of attached images, an area where unimodal systems frequently fail \cite{mukherjee2013,radford2021}.

Visualizing Model Performance

\begin{figure}
    \centering
    \includegraphics[width=3.5in]{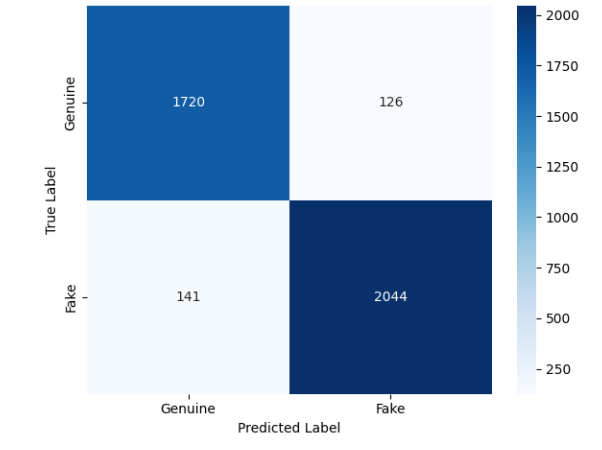}
    \caption{Confusion matrix showing the distribution of predictions for the fake and genuine review classes. The model demonstrates balanced classification performance, with minimal misclassifications, indicating strong discriminative capability across both categories. }
    \label{fig:confusion_matrix}
\end{figure}

To provide deeper insights, a confusion matrix (see Fig.~\ref{fig:confusion_matrix}) was constructed, demonstrating balanced performance across both real and fake review classes. Misclassifications were minimal and typically occurred in borderline cases involving vague textual descriptions or generic stock images, emphasizing the effectiveness of multimodal features in disambiguating such instances.
 \cite{radford2021}.

Additionally, a bar chart illustrating \textbf{ }F1 scores across different models further emphasizes the superiority of the multimodal architecture. This chart clearly shows that while BERT and ResNet individually capture necessary signals, their combination unlocks more discriminative power by fusing semantic and visual contexts.

Significance of Results

These findings are consistent with recent work in multimodal learning where late fusion or attention-based models significantly outperform early fusion or unimodal approaches \cite{mukherjee2013,chen2020}. Moreover, the ability of the model to maintain high F1 scores across diverse review formats ranging from short promotional blurbs to long descriptive feedback demonstrates its adaptability. Such characteristics are crucial for real-world deployment in dynamic environments like e-commerce or travel review platforms \cite{chen2020}.

Furthermore, the relatively low variance in validation accuracy across epochs suggests that the model did not rely on overfitting but learned stable and transferable representations \cite{devlin2019,chen2020}. This positions it as a reliable candidate for broader deployment and further fine-tuning on industry-specific datasets.

\section{Conclusion and Future Work}

This research presents a novel multimodal deep learning framework that integrates BERT for textual representation and ResNet-50 for visual analysis to effectively detect fake reviews across social media platforms. The proliferation of deceptive content online poses significant risks to consumer trust, product reputation, and the credibility of digital marketplaces \cite{ott2011}. By combining the contextual richness of BERT \cite{devlin2019} with the deep visual representation capabilities of ResNet-50 \cite{he2016}, the proposed framework significantly outperforms unimodal baselines, achieving an accuracy of 93.4\% and an F1-score of 0.934. This synergy enables the detection of subtle inconsistencies, such as exaggerated sentiment paired with irrelevant imagery, often indicative of deceptive or manipulative reviews \cite{mukherjee2013}. The model surpasses traditional architectures based on CNNs \cite{kim2014}, LSTMs, and previous multimodal systems \cite{jindal2008,mukherjee2013}, and demonstrates strong generalization across training and validation datasets, indicating robustness and potential for real-world deployment in domains such as e-commerce, food delivery, and hospitality. Beyond its current design, the framework offers extensibility to more advanced architectures such as Vision Transformers (ViT) \cite{khan2022transformers} and CLIP \cite{li2021supervision}, which may enhance joint feature representation and improve classification accuracy across complex multimodal inputs \cite{radford2021}. Contrastive learning strategies may further refine decision boundaries by explicitly modeling similarities and dissimilarities in multimodal data \cite{chen2020}. While the curated dataset provides a representative benchmark, its scope could be expanded to incorporate multilingual content, region-specific nuances, and varying review formats, including video reviews and voice notes, to capture broader user behavior across diverse geographies and platforms. Future directions include the development of real-time, low-latency detection models suitable for deployment on edge devices and content moderation pipelines. Integration of explainability frameworks such as SHAP and LIME \cite{lundberg2017} can enhance interpretability, stakeholder trust, and compliance with emerging transparency regulations. Moreover, ethical considerations surrounding multimodal surveillance, bias in content moderation, and fairness across linguistic and cultural subgroups warrant further investigation. Altogether, this work establishes a scalable, interpretable, and ethically aware foundation for multimodal fake review detection. By leveraging synergistic learning across text and visual modalities, it contributes meaningfully to the broader efforts of preserving digital trust, combating misinformation, and ensuring platform accountability in an increasingly user-generated content ecosystem.

\section{Acknowledgment}
The authors thank all contributors who supported this research through insights, tools, or resources. Special thanks are extended to the open-source community for providing datasets and pre-trained models that were essential for the implementation of the hybrid churn prediction framework. Gratitude is also expressed to the reviewers and advisors for their valuable feedback throughout the development of this work. The source code is available at \url{https://github.com/Suhasnadh/multimodal-fake-reviews-detection}.

\bibliographystyle{IEEEtran}
\bibliography{main}

\end{document}